# UoT-UWF-PartAI at SemEval-2021 Task 5: Self Attention Based Bi-GRU with Multi-Embedding Representation for Toxicity Highlighter

**Hamed Babaei Giglou**
Department of Computer Science
University of Tabriz, Tabriz, Iran
`h.babaei98@ms.tabrizu.ac.ir`

**Taher Rahgooy**
Department of Computer Science
University of West Florida, Florida, USA
`trahgooy@students.uwf.edu`

**Mostafa Rahgouy**
Part AI Research Center
Tehran, Iran
`mostafa.rahgouy@partdp.ai`

**Jafar Razmara**
Department of Computer Science
University of Tabriz, Tabriz, Iran
`razmara@tabrizu.ac.ir`

## Abstract

Toxic Spans Detection(TSD) task is defined as highlighting spans that make a text toxic. Many works have been done to classify a given comment or document as toxic or non-toxic. However, none of those proposed models work at the token level. In this paper, we propose a self-attention-based bidirectional gated recurrent unit(BiGRU) with a multi-embedding representation of the tokens. Our proposed model enriches the representation by a combination of GPT-2, GloVe, and RoBERTa embeddings, which led to promising results. Experimental results show that our proposed approach is very effective in detecting span tokens.

## 1 Introduction

With the massive increase in social interactions on online social networks, keeping discussions fruitful is a central concern for platform providers. Indeed, abusive (e.g., bullying, profanity, hate speech), damaging the reputation of a platform. Thus, it is necessary to be detected by automated machine learning systems because of huge amount of data. However, the previous works mostly focus on whether the given document(e.g. comment) is toxic or not. Detecting the span tokens in the document may be more beneficial. For example, it can prevent users to use span tokens before they send their post. Also, it is useful to filter span tokens before learning AI chatterbots instead of removing the whole documents.

In this paper, we propose a self-attention-based bidirectional gated recurrent unit (BiGRU) with a multi-embedding representation of the tokens. Our proposed model enriches representation showed very promising results.

The rest of the paper is organized as follows. Section 2 provides background and presents some related works on TSD in general. Section 3 introduces our BiGRUs model. Results are covered in Section 4. In Section 5 we draw a conclusion.

## 2 Related Research

A toxic post (comments) is defined as a rude, disrespectful, or unreasonable comment that is likely to make other users leave a discussion. A goal of toxic comment classification is to give a right to freedom of expression on the web.

Training complex neural networks (NN) requires enough datasets of toxic comments. Word embeddings are the basis of the NNs when working with text data. NNs for toxic comment classification use Recurrent Neural Networks (RNN) layers such as Long short-term memory (LSTM) (Hochreiter and Schmidhuber, 1997) or Gated Recurrent Unit (GRU) (Chung et al., 2014) layers. Similarly, the attention mechanism for NNs has been successfully applied to toxic comment classification (Pavlopoulos et al., 2017; van Aken et al., 2018). In semi-automated content-moderation, attention can be considered as a highlighter of abusive or toxic words.

Badjatiya et al. (2017) proposed a system for a hate-speech task based on deep-learning with a combination of LSTM, random embedding, and gradient boosted decision trees as the best model. They used random embedding, GloVe (Pennington et al., 2014), and FastText (Bojanowski et al., 2016) representation for experimentation. They concluded that a combination of CNN and LSTM with FastText or GloVe embedding as features for gradient boosted decision trees can not yield better results.

van Aken et al. (2018) held an in-depth error analysis, and based on their comparance with different deep learning and shallow approaches; they observed three common challenges: out-of-vocabulary words, long-range dependencies, and multi-word phrases. They experimented with various deep learning models, including CNN, LSTM, BiLSTM, GRU, BiGRU, and Attention mechanism with GloVe and FastText embeddings to tackle these challenges. Their experimentation on two datasets showed that BiGRU with Attention mechanism with GloVe and FastText representations achieved promising results with respect to other models. However, in the final, they proposed an ensemble approach that outperforms all individual models.

In (Pavlopoulos et al., 2017) they proposed a deep classification-specific attention mechanism with BiGRU to highlight suspicious words for automatic and semi-automatic content moderations.

## 3 Proposed Method

In this section, we describe the details of our proposed neural network model. Our proposed approach aims to predict whatever the tokens of given comments are toxic or not. Figure 1 depicts an overview of our proposed method.

At first, in a dataset, original posts are tokenized and preprocessed. Each token at each post is labeled as a toxic or not toxic token by their spans. Then, a multi-embedding representation of tokens is created. Next, the Bidirectional GRUs (BiGRUs) models are applied to extract the higher-level feature sequences with sequential information from multi-embeddings. After that, a self-attention mechanism computes attention weight between each pair of elements in a single sequence. Finally, the generated output feature sequences from self-attention-based BiGRUs are fed into the fully-connected dense layer and then into the final prediction module to determine the prediction. In the following, we describe each component elaborately.

### 3.1 Preprocessing

First, each comment is tokenized into words with their spans. Next, tokens are preprocessed; the preprocessing consists of lowercasing and removing punctuations, special characters, numbers, Unicodes, smileys, and emojis. After preprocessing, empty tokens are removed (tokens which empty spaces). Finally, for a post, toxic or not-toxic labels based on grand truth assigned to preprocessed tokens. In total, we obtained **21790** unique words. The obtained words are used as a vocabulary. In the next step, a multi-embedding is used to extract

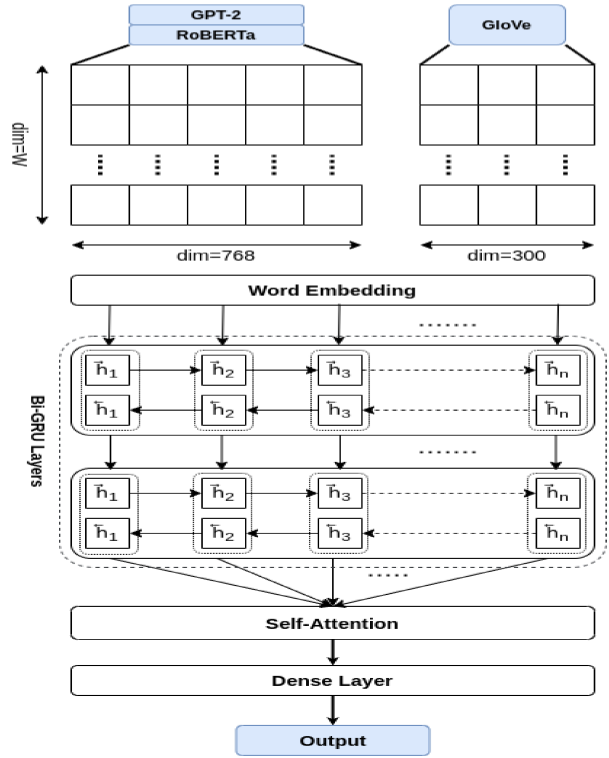

Figure 1: Architecture of Proposed Model

features for unique words to create an embedding matrix for modeling.

### 3.2 Multi-Embedding Layer

Toxic comments often use obfuscations, for example, "f**k u", "Son of a B****", "***k them.". Also, misspelled and abbreviation words are common in online discussions. Word2Vec (Mikolov et al., 2013) and GloVe fail to find a good representation of these words because words never occurred in training time. These words are out-of-vocabulary (OOV) (Risch and Krestel, 2020). However, we can take advantage of failing representations to represent toxic tokens that are not in vocabulary by setting their representations to zero. Regarding this hypothesis, if GloVe or Word2Vec contains OOV, we can set their token representations to zero and use language model embeddings to find subset word representation. By concatenating these two types of representation for each token in the sequence, we can build awareness representation of sequences.

For this purpose, we used the multi-embedding representation of tokens.

**GloVe** is a global log-bilinear regression model with a weighted least-square objective that combines the advantages of global matrix factorization and local context windows. It leverages statistical information by training only on the nonzero elements in a word-word co-occurrence matrix in a large corpus.

**GPT-2** (Radford et al., 2019) is an unsupervised transformer language model for general-purpose learners. It is trained on WebText, which contains over 8 million documents.

**RoBERTa** (Liu et al., 2019) is an optimized version of BERT (Devlin et al., 2018) model. It builds on BERT's language masking strategy. RoBERTa modifies key hyperparameters in BERT, including removing BERT's next sentence pretraining objective and training with much larger mini-batches and learning rates.

To empower representation, we combined RoBERTa and GPT-2 representations (by summing) and then concatenated them with GloVe (840B tokens, 2.2 vocab). We achieved representation matrix with $W \times 1068$ dimension for training vocabs, where $W$ is the number of vocabs in training. During analysis, we found nearly $6k$ OOV words in training, which GloVe does not produce a representation for them. It is a significant number regarding the training vocabulary size.

### 3.3 Bidirectional GRUs

Bidirectional Gated Recurrent Unit (BiGRU) in the central part of Figure 1 is a bidirectional version of GRU. The GRU allows to adaptively capture dependencies from large sequences of data without discarding information from earlier parts of the sequence. BiGRU combines the forward hidden layer with the backward hidden layer, which can process each sequence in both left-to-right and right-to-left order to embed the sequential dependencies in both directions. The first BiGRU layer is used to process each sequence token-by-token and produce an intermediate representation. Then, this intermediate representation is used as input for the second BiGRU layer.

### 3.4 Self-Attention Layer

The words in sequences sometimes are related to each other, like "Son", "of", "B***" and sometimes are not related. To determine how two tokens are related, Self-Attention Networks (SANs) (Luong et al., 2015) produce the output with the same size as input sequences by considering the attention of all input tokens with each other. It learns the important interactions between tokens.

### 3.5 Dense Layer

The dense layer or feed-forward layer is the most general-purpose deep learning layer. The dense layer consists of 50 neurons for the weighted linear combination of inputs with the activation function of tanh to squashes the input to the range [0, 1].

### 3.6 Prediction Module

The final layer of the network has three neurons, and its returned value is a continuous numerical value. We used the sigmoid activation function to produce a probability vector. For loss function and optimization, we employ Sparse Categorical Cross entropy and RMSprop, respectively.

## 4 Results

### 4.1 Dataset

For toxic span detection tasks (Pavlopoulos et al., 2021) posts from publicly available Civil Comment dataset are used for annotations of particular toxic spans in toxic comments. The task consists of 7939 annotated comments with their toxic spans for training and 2000 for the test. However, we treat 690 samples of trial data as a development set for our investigations.

### 4.2 Evaluation

For evaluation of participating systems in the challenge, F1 score presented in (Da San Martino et al., 2019) was used. If we consider $A_i$ to return a set $S_{A_i}^t$ of character offsets for the part of the post found to be toxic, and similarly $G^t$ be the character offset of the grand truth annotation of $t$. We can compute F1 score of system $A_i$ with respect to the $G$ for post $t$ as follows:

$$F_1^t(A_i, G) = \frac{2 \cdot P^t(A_i, G) \cdot R^t(A_i, G)}{P^t(A_i, G) + R^t(A_i, G)}$$

$$P^t(A_i, G) = \frac{|S_{A_i}^t \cap S_G^t|}{S_{A_i}^t}$$

$$R^t(A_i, G) = \frac{|S_{A_i}^t \cup S_G^t|}{S_{A_i}^t}$$

If $S_G^t$ and $S_{A_i}^t$ are empty for some post $t$, then $F1 = 1$, and otherwise $F1 = 0$. In final, to obtain a single score for system $A_i$, the $F1$s averaged over all the posts $t$ of the evaluation dataset.

| Method | GloVe | GPT-2 | RoBERTa | RG | GoR | GoG | Ensemble |
|---|---|---|---|---|---|---|---|
| Results on dev set | | | | | | | |
| BiLSTM | 0.619 | 0.580 | 0.634 | 0.647 | 0.627 | 0.621 | **0.655** |
| BiGRU | 0.597 | 0.641 | 0.621 | 0.664 | 0.637 | **0.668** | 0.643 |
| BiLSTM + Attention | 0.581 | 0.615 | 0.620 | 0.638 | 0.607 | 0.445 | **0.663** |
| BiGRU + Attention | 0.572 | 0.649 | 0.562 | 0.521 | 0.664 | 0.601 | **0.668** |
| Results on Test set | | | | | | | |
| BiLSTM | 0.627 | 0.666 | 0.663 | 0.669 | 0.665 | **0.680** | 0.673 |
| BiGRU | 0.633 | 0.623 | 0.660 | 0.662 | 0.648 | 0.670 | **0.680** |
| BiLSTM + Attention | 0.653 | **0.676** | 0.657 | 0.600 | 0.668 | 0.559 | 0.633 |
| BiGRU + Attention | 0.639 | 0.659 | 0.644 | 0.627 | 0.640 | **0.678** | 0.677 |

Table 1: Experimental Results on Trial (dev) and Test sets. RG refers to the ensemble of RoBERTa and GPT-2 embeddings. Similarly, GoR, refers to the ensemble of GloVe and RoBERTa embeddings, and GoG refers to the ensemble of GloVe and GPT-2 embeddings

### 4.3 Results

For all experimentation, we used Google Colab free GPU[1] to train our models with 10 epochs. We set the batch size to 32, and we pad the comments to the 215 sequence length. We obtained 6330 OOV out of 21790 words in the vocabulary, which GloVe does not produce a representation for them. For experimentation, we used BiLSTM and BiGRU models with SAN followed by a dense layer. Also, we examined representation combinations of GloVe, GPT2, and RoBERTa and reported them in Table 1 for dev (trial) and test sets. According to the experimentations, all models perform well when multi-embedding representation is utilized.

In the first part, we took GloVe representation as our baseline representation. Regarding this representation, in most cases, GPT-2 and RoBERTa perform well (6 cases for GPT-2, and 7 cases for RoBERTa). It shows how much the contextualized representations are useful; however, it is hard to tell among GPT-2 and RoBERTa which one is performing well.

In the second part, we combined different representations that achieved a higher averaged value of F1 score in all cases. Except for one case, namely BiLSTM + Attention, the differences between representation by GPT-2 and GoR is 0.008. In general, an ensemble of embeddings achieved a higher score than single representations.

In the final, because of two reasons, we considered the BiGRU + Attention model with multi-embedding representations as the final model for this task. First, It achieved a higher averaged F1 according to the dev set, and the second higher averaged F1 score according to the test set. The second reason is that the margin between the BiGRU+Attention model in the dev and test set was less than the others (0.009). Submitted model to the competition achieved averaged F1 score of **0.677** and place **23** of the competition among 91 teams.

We shared the implementation of the proposed model in GitHub[2] for the research community.

### 5 Conclusion

In this paper, we presented our approach for SemEval-2021 Task 5: Toxic Span Detection. We tried to tackle the problem by employing multi-embedding and deep learning techniques. We conducted some experiments using different models. For example, we implemented a BiLSTM model, BiLSTM with SA, BiGRU, and BiGRU with SA, but the model that gave a promising result and relied on BiGRU with SA model.

---

[1] https://colab.research.google.com

[2] https://github.com/HamedBabaei/semeval2021-task5-tsd